\documentclass[sigconf, nonacm]{acmart}

\AtBeginDocument{%
  \providecommand\BibTeX{{%
    \normalfont B\kern-0.5em{\scshape i\kern-0.25em b}\kern-0.8em\TeX}}}

\usepackage{amsfonts}
\usepackage{multirow}
\usepackage{amsmath, bm}
\usepackage{subcaption}
\usepackage{bbm}
\usepackage[linesnumbered,ruled,vlined]{algorithm2e}
\SetKwInput{KwInput}{Input}                
\SetKwInput{KwOutput}{Output}  

\begin{document}
\fancyhead{}

\title{Towards Robust Cross-domain Image Understanding with Unsupervised Noise Removal}


\author{Lei Zhu}
\affiliation{%
  \institution{National University of Singapore}
  \country{Singapore}
  }
\email{zhu-lei@comp.nus.edu.sg}

\author{Zhaojing Luo}
\authornote{Corresponding author.}
\affiliation{%
  \institution{National University of Singapore}
  \country{Singapore}
 }
\email{zhaojing@comp.nus.edu.sg}

\author{Wei Wang}
\affiliation{%
  \institution{National University of Singapore}
  \country{Singapore}
}
\email{wangwei@comp.nus.edu.sg}

\author{Meihui Zhang}
\affiliation{%
  \institution{Beijing Institute of Technology}
  \country{China}
}
\email{meihui\_zhang@bit.edu.cn}

\author{Gang Chen}
\affiliation{%
  \institution{Zhejing University}
  \country{China}
}
\email{cg@zju.edu.cn}

\author{Kaiping Zheng}
\affiliation{%
  \institution{National University of Singapore}
  \country{Singapore}
}
\email{kaiping@comp.nus.edu.sg}


\begin{abstract}
Deep learning has made a tremendous impact on various applications in multimedia, such as media interpretation and multimodal retrieval. However, deep learning models usually require a large amount of labeled data to achieve satisfactory performance. In multimedia analysis, domain adaptation studies the problem of cross-domain knowledge transfer from a label rich source domain to a label scarce target domain, thus potentially alleviates the annotation requirement for deep learning models. However, we find that contemporary domain adaptation methods for cross-domain image understanding perform poorly when source domain is noisy. Weakly Supervised Domain Adaptation (WSDA) studies the domain adaptation problem under the scenario where source data can be noisy. Prior methods on WSDA remove noisy source data and align the marginal distribution across domains without considering the fine-grained semantic structure in the embedding space, which have the problem of class misalignment, e.g., features of cats in the target domain might be mapped near features of dogs in the source domain. In this paper, we propose a novel method, termed Noise Tolerant Domain Adaptation (NTDA), for WSDA. Specifically, we adopt the cluster assumption and learn cluster discriminatively with class prototypes (centroids) in the embedding space. We propose to leverage the location information of the data points in the embedding space and model the location information with a Gaussian mixture model to identify noisy source data. We then design a network which incorporates the Gaussian mixture noise model as a sub-module for unsupervised noise removal and propose a novel cluster-level adversarial adaptation method based on the Generative Adversarial Network (GAN) framework which aligns unlabeled target data with the less noisy class prototypes for mapping the semantic structure across domains. Finally, we devise a simple and effective algorithm to train the network from end to end. We conduct extensive experiments to evaluate the effectiveness of our method on both general images and medical images from COVID-19 and e-commerce datasets. The results show that our method significantly outperforms state-of-the-art WSDA methods.
\end{abstract}



\maketitle

\section{Introduction}
There is great interest in using deep learning for various multimedia applications, such as media interpretation~\cite{forgione2018implementation, vukotic2016multimodal}, multimodal retrieval~\cite{wu2019online, liao2018interpretable, wang2016effective, wang2014effective}. Much of its success is attributed to the availability of large-scale labeled training data~\cite{deng2009imagenet}. However, in practice, large-scale labeled data are hardly available, as manual annotating sufficient label information for various multimedia applications is both expensive and time-consuming. Thus, it is desirable to reuse labeled data from a related domain for cross-domain image understanding. This process is called Domain Adaptation (DA), which transfers knowledge from a label rich source domain to a label scarce target domain~\cite{pan2009survey}. Intuitively, the data quality in the source domain affects the domain adaptation performance. However, in practice, high-quality source data related to a target task of interest is hardly available. In contrast, the Internet and social media contain large-scale labeled multimedia data which can be downloaded with keyword search~\cite{xiao2015learning, krause2016unreasonable}, but unfortunately, these data contain noise, either in features, labels or both. Similarly in medical image analysis, annotating medical data requires medical expertise and due to subjectivity of domain experts and diagnostic difficulties, noisy labels are often inevitable. Thus, it is meaningful to study robust domain adaptation under the scenario when source data is noisy in order for better cross-domain image understanding. This problem has been referred to as Weakly Supervised Domain Adaptation (WSDA)~\cite{shu2019transferable}. 

Although WSDA enables many practical use cases of domain adaptation in real life and can substantially reduce the annotation costs, it is still not well studied in the literature. There are two entangled challenges in WSDA, namely source data noise and distribution shift across domains. Directly applying existing domain adaptation methods for WSDA will not work, as source data noise severely deteriorates the adaptation performance~\cite{shu2019transferable, liu2019butterfly}. \citeauthor{liu2019butterfly}~\cite{liu2019butterfly} recently propose a Butterfly framework to address these issues. However, their work only considers the scenario where source domain contains label noise data and cannot handle feature noise data. Another limitation of the method is its large model size, and as a result, it incurs a large amount of computational resources to train the model.

\begin{figure}
\centering
\includegraphics[width=0.6\linewidth]{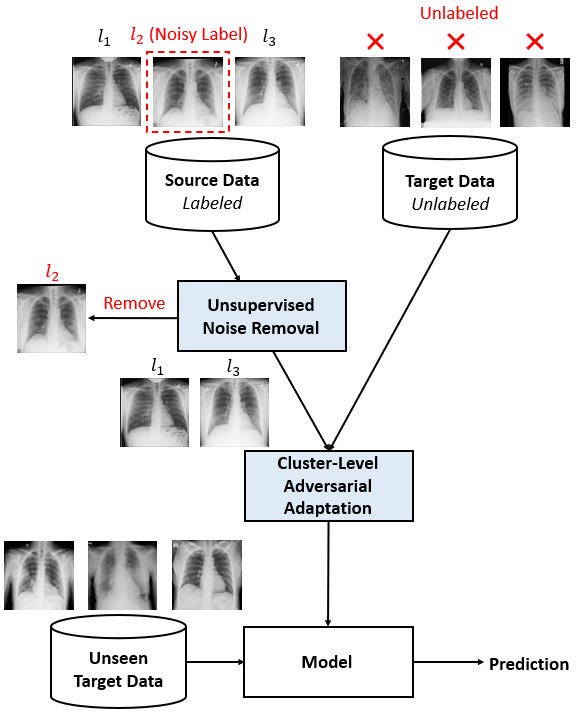}
\caption{Workflow of Noise Tolerant Domain Adaptation for medical image analysis. Source and Target data are chest X-ray images acquired with different scanners for lung disease diagnosis.}
\label{fig: ntda}
\end{figure}

\citeauthor{shu2019transferable}~\cite{shu2019transferable} recently propose transferable curriculum for WSDA. The transferable curriculum can select transferable and clean source data for adversarial domain adaptation, which makes their method robust to source data noise. However, one problem with their method is that the adversarial learning between the feature extractor and domain discriminator only aligns the marginal distribution across domains and ignores the fine-grained class structure in the embedding space. Noisy source data distorts the original source data distribution, thus directly aligning the marginal distribution across domains would potentially transfer the noise information from source domain to target domain and possibly cause class misalignment due to label noise, where even with perfect alignment of the marginal distribution, the class structure across domains may not be well aligned, e.g., features of cats in target domain might be mapped near features of dogs in source domain, which leads to poor target performance~\cite{wang2020classes,combes2020domain}.

Recently, several Unsupervised Domain Adaptation (UDA) methods~\cite{shu2018dirt, deng2019cluster} adopt the cluster assumption~\cite{chapelle2005semi} to alleviate the class misalignment problem, where they assume data distributes in the embedding space with separated data clusters and data samples in the same cluster share the same class label. In~\cite{shu2018dirt}, \citeauthor{shu2018dirt} propose Virtual Adversarial Domain Adaptation (VADA) which combines virtual adversarial training~\cite{miyato2018virtual} and conditional entropy loss to push the decision boundaries away from class clusters in the embedding space. In~\cite{deng2019cluster}, \citeauthor{deng2019cluster} propose Cluster Alignment with a Teacher (CAT), which forces features of both source domain and target domain to form discriminative class-conditional clusters and aligns the corresponding clusters across domains. Although both works alleviate the class misalignment problem and demonstrate significant improvement in performance over prior domain adaptation methods that only align marginal distribution across domains, however, like most existing domain adaptation methods, they perform poorly when source domain contains noise~\cite{shu2019transferable, liu2019butterfly}. 

In light of the issues with existing WSDA methods and inspired from recent clustering based domain adaptation methods, in this paper, we propose a novel method for WSDA with unsupervised noise removal to address these issues. For its noise tolerance property, we call our method, Noise Tolerant Domain Adaptation, or NTDA in short. Specifically, we learn the clusters discriminatively with class prototypes (centroids) in the embedding space. We propose to estimate the probability of a source data point being noisy with a Gaussian mixture noise model based on its distance to the class prototype and filter source data points with high probabilities of being noisy (See Sec.~\ref{sec:3.2} for more details). We incorporate the Gaussian mixture noise model as a sub-module within a deep network and propose a cluster-level adversarial adaptation method based on the GAN framework which aligns unlabeled target data with the less noisy class prototypes for mapping the semantic structure across domains. 
Finally, we devise a simple and effective algorithm to train the network from end to end. 
Fig.~\ref{fig: ntda} presents the workflow of our proposed NTDA method. 

To summarize, we make the following contributions in this paper:
\vspace{-3mm}
\begin{itemize}
    \item We identify several issues with existing WSDA methods and propose a simple and effective method, NTDA, to address these issues.
    \item We propose a novel Unsupervised Noise Removal (UNR) method based on the location information of data points, which can be applied to other clustering based domain adaptation methods to make them robust to source data noise.
    \item We propose a Cluster-Level Adversarial Adaptation (CAA) method, which adversarially aligns target data points with the less noisy class prototypes in the embedding space and alleviates class misalignment problem.
    \item We conduct extensive experiments to evaluate the effectiveness of our method on both general images and medical images. The results show that NTDA significantly improves state-of-the-art results for WSDA.
\end{itemize}
\vspace{-2mm}
NTDA has been developed as part of the library for MLCask~\cite{luo2021mlcask} for supporting healthcare analytics.  MLCask is a model-data provenance and pipeline management system that rides on Apache SINGA~\cite{ooi2015singa} for supporting end-to-end analytics. The  remainder  of  the  paper  is  organized  as  follows. Section~\ref{sec:related} provides a brief background on domain adaption, and related works. In Section~\ref{sec:newwsda}, we present our methodology to tackle WSDA problems and propose NTDA. We conduct extensive experimental study and present the results in Section~\ref{sec:exp}.  We conclude in Section~\ref{sec:conclusions}.

\section{Related Works} \label{sec:related}
\textbf{Domain Adaptation}~\cite{pan2009survey} aims to build models that generalize across domains. Existing domain adaptation methods can be roughly divided into two major categories: discrepancy-based and adversarial-based methods. Discrepancy-based methods align feature distributions across domains by minimizing certain distribution discrepancy, such as Maximum Mean Discrepancy (MMD)~\cite{tzeng2014deep, long2015learning}, correlation distance~\cite{sun2016deep} or Central Moment Discrepancy (CMD)~\cite{zellinger2017central}. Adversarial-based methods draws inspiration from the two-player game of Generative Adversarial Networks (GAN)~\cite{goodfellow2014generative}. DANN~\cite{ganin2014unsupervised} trains domain invariant features via adding a domain classifier in the deep feature learning pipeline via gradient reversal. ADDA~\cite{tzeng2017adversarial} adopts asymmetric feature extractors for adversarial training. CDAN \cite{long2018conditional} conditions the adversarial domain adaptation models on discriminative information conveyed in the classifier prediction. Our method is especially related to the more recent clustering based domain adaptation methods~\cite{shu2018dirt,deng2019cluster,zhang2021prototypical} with cluster assumption.

\textbf{Learning From Noisy Data} is an active research area in machine learning. Recently, \citeauthor{zhang2016understanding}~\cite{zhang2016understanding} empirically demonstrate that noisy data will be memorized by deep networks which destroys their generalization capability. \citeauthor{arpit2017closer}~\cite{arpit2017closer} find that when training with noisy data, deep networks learn simple patterns first before memorizing noisy data. Based on this memorization effect of deep networks, \citeauthor{han2018co}~\cite{han2018co} propose a training paradigm termed ``co-teaching" where they train two networks simultaneously, and utilize the small loss data from one network to teach the other one, which is called the small-loss trick. \citeauthor{yu2019does}~\cite{yu2019does} further propose the ``Update by Disagreement" strategy with ``co-teaching" to prevent the two networks converge to consensus.

\textbf{Weakly Supervised Domain Adaptation} (WSDA) studies the domain adaptation problem under the scenario where source data can be noisy. Although WSDA studies a more practical problem than ordinary domain adaptation, it is still under-explored in the literature. Recently, \citeauthor{liu2019butterfly}~\cite{liu2019butterfly} propose a Butterfly framework which consists of four networks for WSDA. However, their method demands a large amount of computational resources for training. \citeauthor{shu2019transferable}~\cite{shu2019transferable} propose transferable curriculum to select transferable and clean source data for WSDA, but they ignore the class structure in the embedding space for distribution alignment. Yu et al.~\cite{yu2020label} propose a theoretical framework for label-noise robust domain adaptation with a denoising Conditional Invariant Component. However, their method cannot handle cases when there are feature noise in the data. Zhang et al.~\cite{zhang2020collaborative} propose Collaborative Unsupervised Domain Adaptation for general and medical image analysis, where they optimize two networks collaboratively to learn from noisy source data and perform weighted instance-level domain adaptation with unlabeled target data. However, their method aligns the marginal distribution across domains to reduce the distribution shift which suffers from class misalignment under the label noise scenario.

\section{Methodology} \label{sec:newwsda}
In Unsupervised Domain Adaptation (UDA), we are given $N^{s}$ labeled data $\mathbb{D}^s=\{(\bm{x}_i^s, y_i^s)\}_{i=1}^{N^s}$ in source domain and $N^t$ unlabeled data $\mathbb{D}^t=\{\bm{x}_i^t\}_{i=1}^{N^t}$ in target domain. The source and target data share the same set of labels and are sampled from probability distributions $P^s$ and $P^t$ respectively with $P^s \neq P^t$. In Weakly Supervised Domain Adaptation (WSDA), we relax the assumption that $\mathbb{D}^s$ is clean to that $\mathbb{D}^s$ may be corrupted with noise in labels, features or both. The goal of WSDA is to effectively transfer knowledge from noisy source domain to unlabeled target domain.

\subsection{Preliminary: Discriminative Clustering with Class Information}
In this paper, we adopt the cluster assumption~\cite{chapelle2005semi}, where we assume data distribution in the embedding space contains separated data clusters and data samples in the same cluster share the same class label. With labeled source data $\mathbb{D}^s$, we propose to learn the clusters discriminatively with class information. We employ the prototype learning framework from~\cite{yang2018robust} where we assign a prototype for each class in the embedding space. Let $\bm{f}_i^s=F(\bm{x}_i^s; \bm{\Phi})$ be the feature for source data $\bm{x}_i^s$ with label $y_i^s$, $\bm{p}_j$ be the class prototype for the $jth$ class and $\{\bm{p}_j\}_{j=1}^M$ be the set of prototypes, where $F:\mathcal{X} \rightarrow \mathbb{R}^d$ is the feature extractor, $d$ is the embedded feature dimension, $\bm{\Phi}$ is the set of parameters for $F$ and $M$ is the total number of classes. We measure the probability of a data point belonging to a specific class with a softmax over distance to prototypes as shown in Eqn.~\ref{eqn:0}, where $d(\bm{f}, \bm{p}_{j})=||\bm{f} - \bm{p}_{j}||_2^2$ is the squared euclidean distance between two data points in the embedding space and $T$ is the hyper-parameter for scaling the exponent value. To train the network, we employ cross entropy loss on the prediction probability as shown in Eqn.~\ref{eqn:1}.
\begin{align}
    P(y|\bm{f})&=\frac{e^{-\frac{1}{T} d(\bm{f},\bm{p}_{y})}}{\sum_{j=1}^{M}e^{-\frac{1}{T} d(\bm{f},\bm{p}_j)}}, \label{eqn:0} \\
    \mathcal{L}_{cls}(\mathbb{D}^s)&=-\frac{1}{N^s}\sum_{i=1}^{N^s} log(P(y_i^s|\bm{f}_i^s)). \label{eqn:1}
\end{align}

From a perspective of probability, Eqn~\ref{eqn:0} can be viewed as the posterior probability of a data point belonging to a specific class with a mixture of exponential distribution where the prototypes act as the mean representations for each class~\cite{yang2018robust}. Minimizing Eqn.~\ref{eqn:1} increases the posterior probability for each data point belonging to its labeled class. Therefore, data points will cluster around the corresponding class prototypes in the embedding space which conforms the cluster assumption. We further propose a compact regularizer resembling the contrastive loss in~\cite{deng2019cluster}, which minimizes the distances between data points and their class prototypes to make each cluster more compact as follows:
\begin{align}
\mathcal{L}_{reg}(\mathbb{D}^s)=\frac{1}{N^s}\sum_{i=1}^{N^s} d(\bm{f}_i^s, \bm{p}_{y_i^s}). \label{eqn:2}
\end{align}

For prediction, we denote $C(y, \bm{f};\{\bm{p}_j\}_{j=1}^M)=P(y|\bm{f})$ as a distance-based classifier with the class prototypes as its parameters. For a given input, we first obtain its feature with the feature extractor and then we classify it with the category of its nearest prototype, which is also the category with the maximum predicted probability.

\subsection{Unsupervised Noise Removal} \label{sec:3.2}
\begin{figure}[t]
  \includegraphics[width=\linewidth]{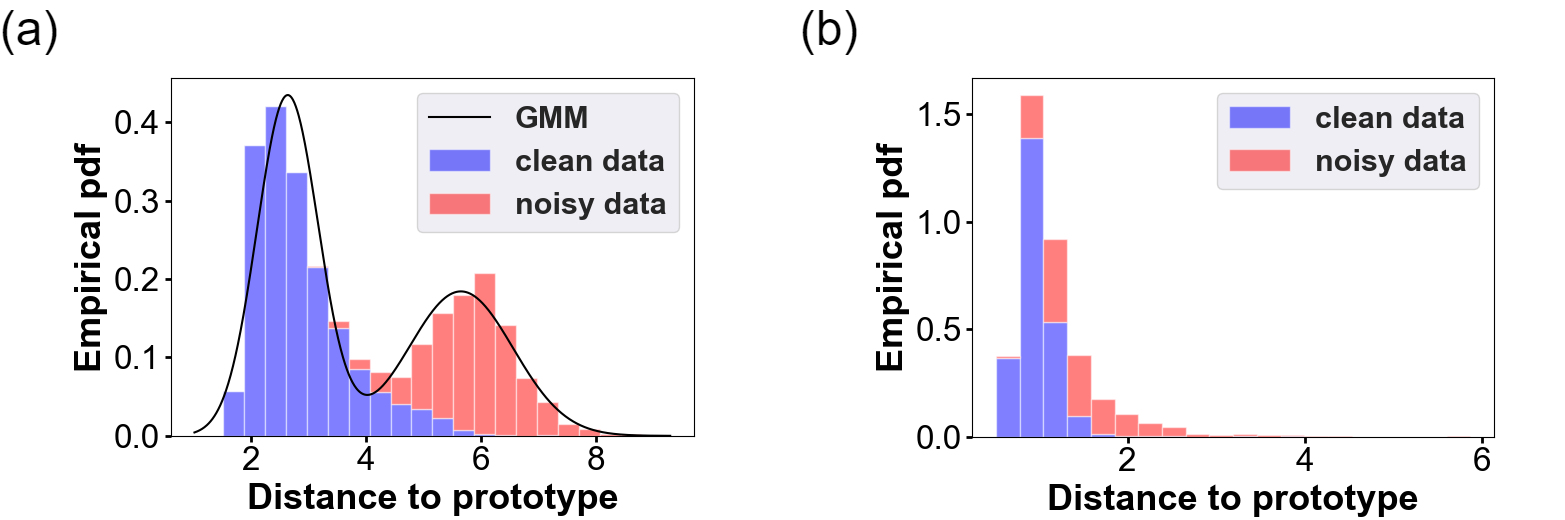}
\caption{Empirical PDF and estimated GMM model on distance to class prototype for 40\% mixed corruption in Amazon dataset after training a network for (a) 10 epochs and (b) 100 epochs.}
\label{fig1}
\end{figure}

We aim to remove noisy source data so that they will not adversely affect domain adaptation~\cite{liu2019butterfly, shu2019transferable}. With discriminative clustering from previous subsection, we observe that noisy source data locate quite differently in the embedding space compared to clean source data at the early training phase of deep networks. We train a deep network with objective $\mathcal{L}_{cls}+0.5*\mathcal{L}_{reg}$ for 100 epochs and measure the distribution of distances for both clean and noisy source data to their class prototypes at (a) epoch 10 (early phase) and (b) epoch 100 (late phase) as shown in Fig.~\ref{fig1}. We observe that clean data locate closer to their class prototypes than almost all noisy data at the early phase of training but some noisy data also locate very close to their class prototypes at the late phase of training.

To understand this phenomenon, we refer to \citeauthor{arpit2017closer}'s study~\cite{arpit2017closer} on the memorization effect of deep networks. \citeauthor{arpit2017closer} find that when training with noisy data, deep networks learn simple patterns first before memorizing noisy data. Our observation conform \citeauthor{arpit2017closer}'s finding. At the early phase of training, as deep networks learn simple patterns first, clean data share simple patterns with each other will form class-wise clusters in the embedding space, thus clean data will locate closer to their class prototypes than noisy data. At the late phase of training, deep networks will memorize the complicated input-to-label patterns from label noise data and the complicated noise patterns from feature noise data, thus, noisy data will also locate close to their class prototypes. 

The disparate distribution of distances to prototypes between clean and noisy source data at the early training phase of deep networks suggests the use of a two-component mixture model~\cite{bishop2006pattern} to estimate the probability of a data point being clean based on its distance to the class prototype. In this paper, we propose a two-component Gaussian mixture model for this purpose as we empirically find it fits both the distance distribution of clean data and noisy data well as shown in Fig.~\ref{fig1}(a). The probability density function of a two-component Gaussian distribution is defined as $p(d)=\sum_{k=1}^2\alpha_kp(d|k)$, where $\alpha_{k}$ is the prior probability for clean ($k=1$) or noisy ($k=2$) data and $p(d|k)=\mathcal{N}(d|\mu_k,\sigma_k)$ is the corresponding normal distance distribution with mean $\mu_k$ and covariance $\sigma_k$. We employ the Expectation-Maximization algorithm~\cite{dempster1977maximum} to estimate the parameters of the Gaussian mixture model and calculate the posterior probability of a data point being clean as follows:
\begin{align}
p(k=1|d)=\frac{\alpha_1\mathcal{N}(d|\mu_1, \sigma_1)}{\sum_{k=1}^2\alpha_k\mathcal{N}(d|\mu_k, \sigma_k)}. \label{eqn:3}
\end{align}
With the unsupervised Gaussian mixture noise model, we will remove data points with large probabilities of being noisy to prevent them affecting domain adaptation.

\subsection{Cluster-Level Adversarial Adaptation}
Noisy source data distort the real source data distribution, which make it more error-prone when aligning the data distribution across domains. More severely, due to label noise, the network cannot easily discriminate two data points from two different classes as they might be annotated with the same label. Thus there is no clear boundary between different classes in the embedding space and data from different classes would mix up with each other, which causes the class misalignment problem even more prominent for domain adaptation. Existing WSDA methods~\cite{shu2019transferable,zhang2020collaborative} which align the marginal distribution across domains however would fail to resolve such issue and potentially transfer noise information from source domain to target domain. To address the problem, our idea is to align the target data with the more reliable class prototypes in the embedding space so that we can reduce the distribution shift across domains.

Specifically, we find that the prediction entropy of a data point encodes the location information of it in the embedding space. If a data point has low prediction entropy, it will locate close to some class prototypes in the embedding space (see Eqn.~\ref{eqn:0}) and if a data point has high prediction entropy, it will locate near the decision boundaries between class prototypes. Source data cluster around their class prototypes in the embedding space, thus they will have low prediction entropy. However, due to the distribution shift across domains, except for some easy-to-transfer target data, which will locate close to some class prototypes in the embedding space like source data, other target data will locate near the decision boundaries~\cite{saito2018maximum, chen2019progressive}. Thus, we design a domain discriminator based on a data point's prediction entropy. The domain discriminator is defined as $D(\bm{f};\{p_j\}_{j=1}^{M})=-\frac{1}{log(M)}\sum_{j=i}^{M}P(j|\bm{f})log(P(j|\bm{f}))$, which calculates the normalized prediction entropy of a data point as the probability of the data point belonging to the target domain, where the $\frac{1}{log(M)}$ term is used to normalize the output within the interval $[0,1]$.

Different from existing adversarial-based domain adaptation methods~\cite{ganin2014unsupervised, tzeng2017adversarial, long2018conditional}, our domain discriminator shares its parameters with the classifier. Since optimizing the classifier with the source data already ensures the prediction entropy on the source data is small, we train the domain discriminator with the cross entropy loss only on target data as follows:
\begin{align}
\begin{split}
\mathcal{L}_{adv_D}(\mathbb{D}^t)=-\frac{1}{N^t}\sum_{i=1}^{N^t} log(D(\bm{f}_i^t)). \label{eqn:4}
\end{split}
\end{align}
Minimizing $\mathcal{L}_{adv_D}$ ensures the domain discriminator $D$ can distinguish target data from source data based on their prediction entropy. To reduce the distribution shift across domains, we adversarially train the feature extractor with the GAN loss~\cite{goodfellow2014generative} as follows:
\begin{align}
\begin{split}
\mathcal{L}_{adv_F}(\mathbb{D}^t)=-\frac{1}{N^t}\sum_{i=1}^{N^t} log(1-D(\bm{f}_i^t)). \label{eqn:5}
\end{split}
\end{align}

Minimizing $\mathcal{L}_{adv_F}$ with the feature extractor will align target data towards their corresponding class prototypes in the embedding space to decrease their prediction entropy for confusing the domain discriminator. As class prototypes are the representative of each class in the embedding space, they are generally more reliable and less noisy compared to source data points, thus, aligning target data towards the less noisy class prototypes is beneficial in the WSDA scenario. In addition, our adaptation method maps target data towards the source data clusters at cluster level with consideration of source data's semantic structure, thus our method alleviates the class misalignment problem. 

\begin{figure}[t]
\begin{center}
\includegraphics[width=\linewidth]{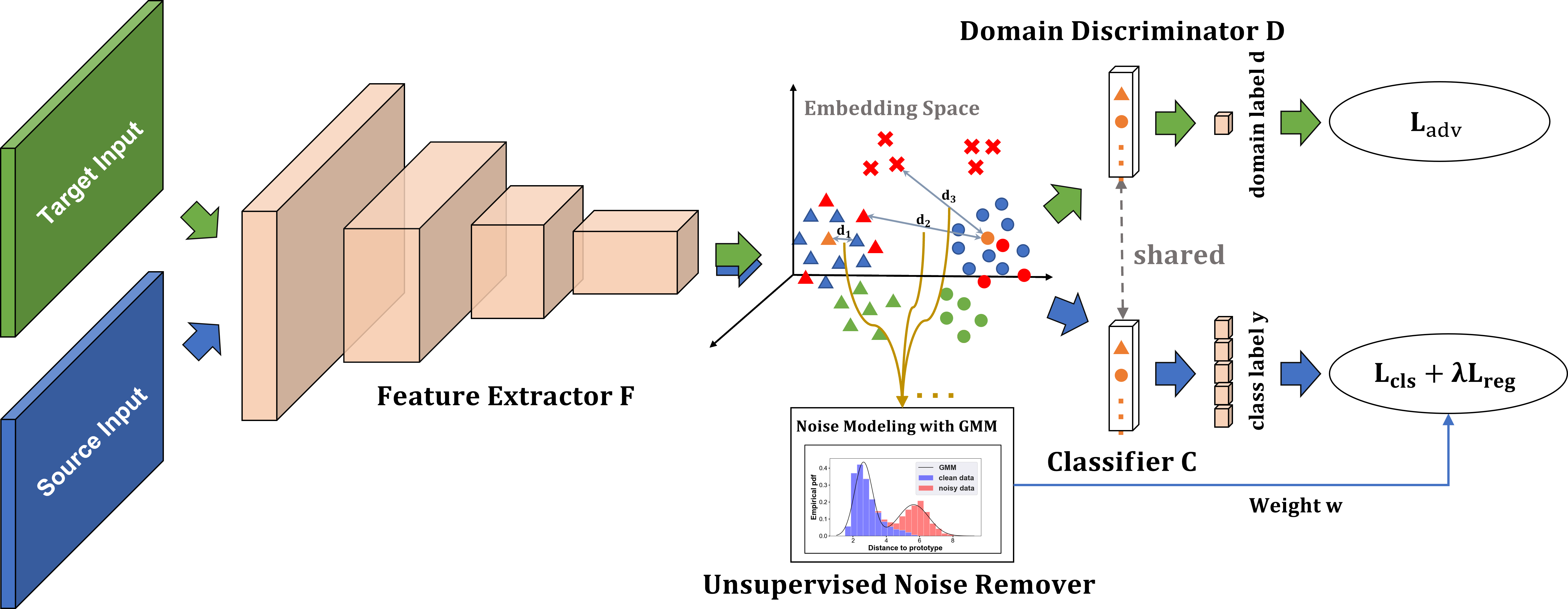}
\end{center}
   \caption{The overview of NTDA network architecture for WSDA with unsupervised noise removal and cluster-level adversarial adaptation. The network consists of a feature extractor, a domain discriminator, a label classifier and an unsupervised noise remover. Blue color represents the source domain, green color represents the target domain. In the embedding space, the red color represents noisy source data. Different shapes in the embedding space represents different classes and the ``X" shape represents feature noise data.
   }
\label{fig2}
\end{figure}

\subsection{Optimization}
In previous subsections, we have introduced different components of our network. In this subsection, we will combine these components into our final network and provide a simple and effective algorithm to train the network from end to end. First, to remove noisy source data, we propose a weighting scheme based on the Gaussian mixture noise model. Denote $d_i^s=\sqrt{d(\bm{f}_i^s, \bm{p}_{y_i^s})}$ as the euclidean distance of source data $\bm{x}_i^s$ to its class prototype $\bm{p}_{y_i^s}$ in the embedding space, the weight for $\bm{x}_i^s$ is defined as follows:
\begin{align}
w(\bm{x}_i^s)= \mathbbm{1}(p(1|d_i^s)>\eta)\frac{p(1|d_i^s)-\eta}{1-\eta}, \label{eqn:6}
\end{align}
where $\mathbbm{1}(\cdot)$ is the indicator function, i.e., it returns 1 when the condition inside the brackets is true and returns 0 otherwise and $\eta$ is the threshold hyper-parameter within interval $[0,1]$. 
We weight the supervision loss for source data as follows:
\begin{align}
    \mathcal{L}_{cls_w}(\mathbb{D}^s)&=-\frac{1}{N^s}\sum_{i=1}^{N^s}w(\bm{x}_i^s)log(P(y_i^s|\bm{f}_i^s)), \label{eqn:7} \\
    \mathcal{L}_{reg_w}(\mathbb{D}^s)&=\frac{1}{N^s}\sum_{i=1}^{N^s}w(\bm{x}_i^s)d(\bm{f}_i^s, \bm{p}_{y_i^s}). \label{eqn:8}
\end{align}
The weighting scheme only selects source data points whose probability of being clean is larger than $\eta$ for training and it linearly scales the probability as the weight for the selected source data so that source data with higher probability of being clean will have larger weight. We present the network architecture of our model in Fig.~\ref{fig2} and the overall objective function to train the network is defined as follows:
\begin{align}
     &\min_{\{\bm{p}_j\}_{j=1}^M} \mathcal{L}_{cls_w}(\mathbb{D}^s)+\lambda_1\mathcal{L}_{reg_w}(\mathbb{D}^s)+\lambda_2 \mathcal{L}_{adv_D}(\mathbb{D}^t), \label{eqn:9} \\
     &\quad \min_{\Phi} \, \mathcal{L}_{cls_w}(\mathbb{D}^s)+\lambda_1\mathcal{L}_{reg_w}(\mathbb{D}^s)+\lambda_2 \mathcal{L}_{adv_F}(\mathbb{D}^t), \label{eqn:10}
\end{align}
where $\lambda_1$ and $\lambda_2$ are two trade-off hyper-parameters.

We present the algorithm to train the network in Alg.~\ref{alg:1}. 
Our algorithm warms up the network for $N_p$ epochs to ensure the network learns some simple patterns first for unsupervised noise modeling. Note $N_p$ can be chosen optimally via inspecting the distance distribution of source data as shown in Fig.~\ref{fig1}. After warming up, our algorithm alternatively performs unsupervised noise removal and cluster-level adversarial adaptation. As our network trains mostly on clean source data, clean source data will be drawn closer and closer to the class prototypes in the embedding space which will further separate the distance distribution of clean and noisy source data apart and make our unsupervised noise model more accurate for selecting clean source data. 
This positive cycle between the unsupervised noise model and the network greatly boosts the performance of our method.

\begin{algorithm}
  \KwInput{Warm up epoch $N_p$, Training epoch $N_t$, Feature extractor $F$, Classifier $C$, Domain discriminator $D$, Batch size $B$, $T$, $\eta$, $\lambda_1$, $\lambda_2$, }
  \KwOutput{Feature extractor $F$ and Classifier $C$.}
   \For{$N_p$ epoch}
        { 
        	Train $F$ and $C$ on $\mathcal{L}_{cls}(\mathbb{D}^s)+\lambda_1\mathcal{L}_{reg}(\mathbb{D}^s)$ with batch size $B$.
        }
    \For{$N_t$ epoch}
        { 
        	Model the distance distribution with a Gaussian mixture model on source data and calculates weights for source data with Eqn.~\ref{eqn:6}. \\
        	Train $F$, $C$ and $D$ on Eqn.~\ref{eqn:9} and Eqn.~\ref{eqn:10} with batch size $B$ after removing source data with weight 0.
        }
\caption{Training algorithm of NTDA model}
\label{alg:1}
\end{algorithm}

\begin{table*}[ht]
\begin{center}
\resizebox{\linewidth}{!}{%
\begin{tabular}{c|p{0.04\linewidth}p{0.04\linewidth}p{0.04\linewidth}p{0.04\linewidth}p{0.04\linewidth}p{0.04\linewidth}p{0.04\linewidth}|p{0.04\linewidth}p{0.04\linewidth}p{0.04\linewidth}p{0.04\linewidth}p{0.04\linewidth}p{0.04\linewidth}p{0.04\linewidth}|p{0.04\linewidth}p{0.04\linewidth}p{0.04\linewidth}p{0.04\linewidth}p{0.04\linewidth}p{0.04\linewidth}p{0.04\linewidth}}
\hline
\multirow{2}{*}{Method} & \multicolumn{7}{c|}{Label Corruption} & %
    \multicolumn{7}{c|}{Feature Corruption} & \multicolumn{7}{c}{Mixed Corruption} \\
\cline{2-22}
 &A$\rightarrow$W&W$\rightarrow$A&A$\rightarrow$D&D$\rightarrow$A&W$\rightarrow$D&D$\rightarrow$W&Avg&A$\rightarrow$W&W$\rightarrow$A&A$\rightarrow$D&D$\rightarrow$A&W$\rightarrow$D&D$\rightarrow$W&Avg&A$\rightarrow$W&W$\rightarrow$A&A$\rightarrow$D&D$\rightarrow$A&W$\rightarrow$D&D$\rightarrow$W&Avg \\
\hline
ResNet~\cite{he2016deep}&\multicolumn{1}{c}{47.2}&\multicolumn{1}{c}{33.0}&\multicolumn{1}{c}{47.1}&\multicolumn{1}{c}{31.0}&\multicolumn{1}{c}{68.0}&\multicolumn{1}{c}{58.8}&\multicolumn{1}{c}{47.5}&\multicolumn{1}{|c}{70.2}&\multicolumn{1}{c}{55.1}&\multicolumn{1}{c}{73.0}&\multicolumn{1}{c}{55.0}&\multicolumn{1}{c}{94.5}&\multicolumn{1}{c}{87.2}&\multicolumn{1}{c}{72.5}&\multicolumn{1}{|c}{58.8}&\multicolumn{1}{c}{39.1}&\multicolumn{1}{c}{69.3}&\multicolumn{1}{c}{37.7}&\multicolumn{1}{c}{75.2}&\multicolumn{1}{c}{75.5}&\multicolumn{1}{c}{59.3} \\
DANN~\cite{ganin2014unsupervised}&\multicolumn{1}{c}{61.2}&\multicolumn{1}{c}{46.2}&\multicolumn{1}{c}{57.4}&\multicolumn{1}{c}{42.4}&\multicolumn{1}{c}{74.5}&\multicolumn{1}{c}{62.0}&\multicolumn{1}{c}{57.3}&\multicolumn{1}{|c}{71.3}&\multicolumn{1}{c}{54.1}&\multicolumn{1}{c}{69.0}&\multicolumn{1}{c}{54.1}&\multicolumn{1}{c}{84.5}&\multicolumn{1}{c}{84.6}&\multicolumn{1}{c}{69.6}&\multicolumn{1}{|c}{69.7}&\multicolumn{1}{c}{50.0}&\multicolumn{1}{c}{69.5}&\multicolumn{1}{c}{49.1}&\multicolumn{1}{c}{80.1}&\multicolumn{1}{c}{79.7}&\multicolumn{1}{c}{66.4} \\
DAN~\cite{long2015learning}&\multicolumn{1}{c}{63.2}&\multicolumn{1}{c}{39.0}&\multicolumn{1}{c}{58.0}&\multicolumn{1}{c}{36.7}&\multicolumn{1}{c}{71.6}&\multicolumn{1}{c}{61.6}&\multicolumn{1}{c}{55.0}&\multicolumn{1}{|c}{73.9}&\multicolumn{1}{c}{60.2}&\multicolumn{1}{c}{72.2}&\multicolumn{1}{c}{59.6}&\multicolumn{1}{c}{92.5}&\multicolumn{1}{c}{88.0}&\multicolumn{1}{c}{74.4}&\multicolumn{1}{|c}{64.4}&\multicolumn{1}{c}{45.1}&\multicolumn{1}{c}{71.2}&\multicolumn{1}{c}{44.7}&\multicolumn{1}{c}{79.3}&\multicolumn{1}{c}{78.3}&\multicolumn{1}{c}{63.8} \\
RTN~\cite{long2016unsupervised}&\multicolumn{1}{c}{64.6}&\multicolumn{1}{c}{56.2}&\multicolumn{1}{c}{76.1}&\multicolumn{1}{c}{49.0}&\multicolumn{1}{c}{82.7}&\multicolumn{1}{c}{71.7}&\multicolumn{1}{c}{66.7}&\multicolumn{1}{|c}{81.0}&\multicolumn{1}{c}{64.6}&\multicolumn{1}{c}{81.3}&\multicolumn{1}{c}{62.3}&\multicolumn{1}{c}{95.2}&\multicolumn{1}{c}{91.0}&\multicolumn{1}{c}{79.2}&\multicolumn{1}{|c}{76.7}&\multicolumn{1}{c}{56.9}&\multicolumn{1}{c}{84.1}&\multicolumn{1}{c}{56.4}&\multicolumn{1}{c}{93.0}&\multicolumn{1}{c}{86.7}&\multicolumn{1}{c}{75.6} \\
ADDA~\cite{tzeng2017adversarial}&\multicolumn{1}{c}{61.5}&\multicolumn{1}{c}{49.2}&\multicolumn{1}{c}{61.2}&\multicolumn{1}{c}{45.5}&\multicolumn{1}{c}{74.7}&\multicolumn{1}{c}{65.1}&\multicolumn{1}{c}{59.5}&\multicolumn{1}{|c}{76.8}&\multicolumn{1}{c}{62.0}&\multicolumn{1}{c}{79.8}&\multicolumn{1}{c}{60.1}&\multicolumn{1}{c}{93.7}&\multicolumn{1}{c}{89.3}&\multicolumn{1}{c}{77.0}&\multicolumn{1}{|c}{69.7}&\multicolumn{1}{c}{54.5}&\multicolumn{1}{c}{72.4}&\multicolumn{1}{c}{56.0}&\multicolumn{1}{c}{87.5}&\multicolumn{1}{c}{85.5}&\multicolumn{1}{c}{70.9} \\
CDAN+E~\cite{long2018conditional}&\multicolumn{1}{c}{78.1}&\multicolumn{1}{c}{60.0}&\multicolumn{1}{c}{75.5}&\multicolumn{1}{c}{50.6}&\multicolumn{1}{c}{85.1}&\multicolumn{1}{c}{77.0}&\multicolumn{1}{c}{71.1}&\multicolumn{1}{|c}{85.2}&\multicolumn{1}{c}{61.1}&\multicolumn{1}{c}{84.9}&\multicolumn{1}{c}{60.0}&\multicolumn{1}{c}{96.8}&\multicolumn{1}{c}{92.8}&\multicolumn{1}{c}{80.1}&\multicolumn{1}{|c}{85.9}&\multicolumn{1}{c}{60.7}&\multicolumn{1}{c}{85.1}&\multicolumn{1}{c}{67.8}&\multicolumn{1}{c}{95.8}&\multicolumn{1}{c}{93.3}&\multicolumn{1}{c}{81.4} \\
\hline
TCL~\cite{shu2019transferable}&\multicolumn{1}{c}{82.0}&\multicolumn{1}{c}{65.7}&\multicolumn{1}{c}{83.3}&\multicolumn{1}{c}{60.5}&\multicolumn{1}{c}{90.8}&\multicolumn{1}{c}{77.2}&\multicolumn{1}{c}{76.6}&\multicolumn{1}{|c}{84.9}&\multicolumn{1}{c}{62.3}&\multicolumn{1}{c}{83.7}&\multicolumn{1}{c}{64.0}&\multicolumn{1}{c}{93.4}&\multicolumn{1}{c}{91.3}&\multicolumn{1}{c}{79.9}&\multicolumn{1}{|c}{87.4}&\multicolumn{1}{c}{64.6}&\multicolumn{1}{c}{83.1}&\multicolumn{1}{c}{62.2}&\multicolumn{1}{c}{\textbf{99.0}}&\multicolumn{1}{c}{92.7}&\multicolumn{1}{c}{81.5} \\
CoUDA~\cite{zhang2020collaborative}&\multicolumn{1}{c}{85.8}&\multicolumn{1}{c}{63.2}&\multicolumn{1}{c}{85.9}&\multicolumn{1}{c}{59.4}&\multicolumn{1}{c}{87.6}&\multicolumn{1}{c}{80.3}&\multicolumn{1}{c}{77.0}&\multicolumn{1}{|c}{84.7}&\multicolumn{1}{c}{64.2}&\multicolumn{1}{c}{81.2}&\multicolumn{1}{c}{62.5}&\multicolumn{1}{c}{96.7}&\multicolumn{1}{c}{\textbf{93.3}}&\multicolumn{1}{c}{80.4}&\multicolumn{1}{|c}{88.3}&\multicolumn{1}{c}{63.4}&\multicolumn{1}{c}{\textbf{89.7}}&\multicolumn{1}{c}{61.4}&\multicolumn{1}{c}{95.5}&\multicolumn{1}{c}{91.9}&\multicolumn{1}{c}{81.7}\\
\hline
\textbf{NTDA}&\multicolumn{1}{c}{\textbf{86.1}}&\multicolumn{1}{c}{\textbf{66.9}}&\multicolumn{1}{c}{\textbf{87.7}}&\multicolumn{1}{c}{\textbf{67.5}}&\multicolumn{1}{c}{\textbf{99.3}}&\multicolumn{1}{c}{\textbf{95.7}}&\multicolumn{1}{c}{\textbf{83.8}}&\multicolumn{1}{|c}{\textbf{86.0}}&\multicolumn{1}{c}{\textbf{66.7}}&\multicolumn{1}{c}{\textbf{87.8}}&\multicolumn{1}{c}{\textbf{66.6}}&\multicolumn{1}{c}{\textbf{96.9}}&\multicolumn{1}{c}{93.0}&\multicolumn{1}{c}{\textbf{82.8}}&\multicolumn{1}{|c}{\textbf{89.5}}&\multicolumn{1}{c}{\textbf{68.3}}&\multicolumn{1}{c}{87.1}&\multicolumn{1}{c}{\textbf{68.8}}&\multicolumn{1}{c}{98.7}&\multicolumn{1}{c}{\textbf{95.1}}&\multicolumn{1}{c}{\textbf{84.6}}
 \\
\hline
\end{tabular}}
\end{center}
\caption{Classification Accuracy (\%) on \textbf{Office-31} with 40\% corruption of labels and features.}
\label{table:1}
\vspace{-4mm}
\end{table*}

\begin{table*}[ht]
\begin{center}
\resizebox{\linewidth}{!}{%
\begin{tabular}{c|p{0.045\linewidth}p{0.045\linewidth}p{0.045\linewidth}p{0.045\linewidth}p{0.045\linewidth}p{0.045\linewidth}p{0.045\linewidth}p{0.045\linewidth}p{0.045\linewidth}p{0.045\linewidth}p{0.045\linewidth}p{0.045\linewidth}p{0.045\linewidth}|p{0.045\linewidth}}
\hline
\multirow{2}{*}{Method} & \multicolumn{13}{c|}{Office-Home} & \multicolumn{1}{c}{Bing-Caltech} \\
\cline{2-15}
&\multicolumn{1}{c}{Ar$\rightarrow$Cl}&\multicolumn{1}{c}{Ar$\rightarrow$Pr}&\multicolumn{1}{c}{Ar$\rightarrow$Rw}&\multicolumn{1}{c}{Cl$\rightarrow$Ar}&\multicolumn{1}{c}{Cl$\rightarrow$Pr}&\multicolumn{1}{c}{Cl$\rightarrow$Rw}&\multicolumn{1}{c}{Pr$\rightarrow$Ar}&\multicolumn{1}{c}{Pr$\rightarrow$Cl}&\multicolumn{1}{c}{Pr$\rightarrow$Rw}&\multicolumn{1}{c}{Rw$\rightarrow$Ar}&\multicolumn{1}{c}{Rw$\rightarrow$Cl}&\multicolumn{1}{c}{Rw$\rightarrow$Pr}&\multicolumn{1}{c}{Avg}&\multicolumn{1}{|c}{B$\rightarrow$C} \\
\hline
ResNet~\cite{he2016deep}&\multicolumn{1}{c}{27.1}&\multicolumn{1}{c}{50.7}&\multicolumn{1}{c}{61.7}&\multicolumn{1}{c}{41.1}&\multicolumn{1}{c}{53.8}&\multicolumn{1}{c}{56.3}&\multicolumn{1}{c}{40.9}&\multicolumn{1}{c}{28.0}&\multicolumn{1}{c}{61.8}&\multicolumn{1}{c}{51.3}&\multicolumn{1}{c}{33.0}&\multicolumn{1}{c}{65.9}&\multicolumn{1}{c}{47.6}&\multicolumn{1}{|c}{74.4} \\
DANN~\cite{ganin2014unsupervised}&\multicolumn{1}{c}{32.9}&\multicolumn{1}{c}{50.6}&\multicolumn{1}{c}{60.1}&\multicolumn{1}{c}{38.6}&\multicolumn{1}{c}{49.2}&\multicolumn{1}{c}{50.6}&\multicolumn{1}{c}{39.9}&\multicolumn{1}{c}{32.6}&\multicolumn{1}{c}{60.4}&\multicolumn{1}{c}{50.5}&\multicolumn{1}{c}{38.4}&\multicolumn{1}{c}{67.4}&\multicolumn{1}{c}{47.6}&\multicolumn{1}{|c}{72.3} \\
DAN~\cite{long2015learning}&\multicolumn{1}{c}{40.9}&\multicolumn{1}{c}{54.2}&\multicolumn{1}{c}{63.0}&\multicolumn{1}{c}{47.2}&\multicolumn{1}{c}{54.3}&\multicolumn{1}{c}{56.3}&\multicolumn{1}{c}{47.2}&\multicolumn{1}{c}{42.8}&\multicolumn{1}{c}{69.0}&\multicolumn{1}{c}{61.0}&\multicolumn{1}{c}{47.4}&\multicolumn{1}{c}{71.9}&\multicolumn{1}{c}{54.6}&\multicolumn{1}{|c}{75.0} \\
RTN~\cite{long2016unsupervised}&\multicolumn{1}{c}{29.3}&\multicolumn{1}{c}{57.8}&\multicolumn{1}{c}{66.3}&\multicolumn{1}{c}{44.0}&\multicolumn{1}{c}{58.6}&\multicolumn{1}{c}{58.3}&\multicolumn{1}{c}{46.0}&\multicolumn{1}{c}{30.1}&\multicolumn{1}{c}{67.5}&\multicolumn{1}{c}{56.3}&\multicolumn{1}{c}{32.2}&\multicolumn{1}{c}{69.9}&\multicolumn{1}{c}{51.4}&\multicolumn{1}{|c}{75.8} \\
ADDA~\cite{tzeng2017adversarial}&\multicolumn{1}{c}{32.6}&\multicolumn{1}{c}{52.0}&\multicolumn{1}{c}{60.6}&\multicolumn{1}{c}{42.6}&\multicolumn{1}{c}{53.5}&\multicolumn{1}{c}{54.3}&\multicolumn{1}{c}{43.0}&\multicolumn{1}{c}{31.6}&\multicolumn{1}{c}{63.1}&\multicolumn{1}{c}{52.7}&\multicolumn{1}{c}{37.7}&\multicolumn{1}{c}{67.5}&\multicolumn{1}{c}{49.3}&\multicolumn{1}{|c}{74.7} \\
CDAN+E~\cite{long2018conditional}&\multicolumn{1}{c}{41.1}&\multicolumn{1}{c}{61.6}&\multicolumn{1}{c}{69.3}&\multicolumn{1}{c}{49.2}&\multicolumn{1}{c}{65.0}&\multicolumn{1}{c}{63.9}&\multicolumn{1}{c}{47.1}&\multicolumn{1}{c}{41.5}&\multicolumn{1}{c}{70.8}&\multicolumn{1}{c}{61.3}&\multicolumn{1}{c}{45.4}&\multicolumn{1}{c}{76.3}&\multicolumn{1}{c}{57.7}&\multicolumn{1}{|c}{82.6} \\
\hline
TCL~\cite{shu2019transferable}&\multicolumn{1}{c}{38.8}&\multicolumn{1}{c}{62.1}&\multicolumn{1}{c}{69.4}&\multicolumn{1}{c}{46.5}&\multicolumn{1}{c}{58.5}&\multicolumn{1}{c}{59.8}&\multicolumn{1}{c}{51.3}&\multicolumn{1}{c}{39.9}&\multicolumn{1}{c}{72.3}&\multicolumn{1}{c}{63.4}&\multicolumn{1}{c}{43.5}&\multicolumn{1}{c}{74.0}&\multicolumn{1}{c}{56.6}&\multicolumn{1}{|c}{79.0} \\
CoUDA~\cite{zhang2020collaborative}&\multicolumn{1}{c}{39.7}&\multicolumn{1}{c}{62.3}&\multicolumn{1}{c}{68.8}&\multicolumn{1}{c}{52.5}&\multicolumn{1}{c}{61.4}&\multicolumn{1}{c}{63.3}&\multicolumn{1}{c}{47.9}&\multicolumn{1}{c}{42.6}&\multicolumn{1}{c}{70.8}&\multicolumn{1}{c}{63.6}&\multicolumn{1}{c}{49.9}&\multicolumn{1}{c}{75.1}&\multicolumn{1}{c}{58.1}&\multicolumn{1}{|c}{79.1} \\
\hline
\textbf{NTDA}&\multicolumn{1}{c}{\textbf{48.3}}&\multicolumn{1}{c}{\textbf{67.2}}&\multicolumn{1}{c}{\textbf{73.9}}&\multicolumn{1}{c}{\textbf{55.0}}&\multicolumn{1}{c}{\textbf{65.8}}&\multicolumn{1}{c}{\textbf{64.8}}&\multicolumn{1}{c}{\textbf{57.8}}&\multicolumn{1}{c}{\textbf{48.9}}&\multicolumn{1}{c}{\textbf{76.6}}&\multicolumn{1}{c}{\textbf{68.1}}&\multicolumn{1}{c}{\textbf{54.0}}&\multicolumn{1}{c}{\textbf{79.3}}&\multicolumn{1}{c}{\textbf{63.3}}&\multicolumn{1}{|c}{\textbf{83.9}} \\
\hline
\end{tabular}}
\end{center}
\caption{Classification Accuracy (\%) on \textbf{Office-Home} with 40\% mixed corruption and \textbf{Bing-Caltech} with native noise.}
\label{table:2}
\vspace{-4mm}
\end{table*}

\section{Experiments} \label{sec:exp}
\subsection{Datasets and Experimental Settings}
\textbf{Office-31}~\cite{saenko2010adapting} is a benchmark dataset for domain adaptation, consisting of 4652 images with 31 classes in 3 distinct domains: Amazon (A), Webcam (W), DSLR (D). By permuting the 3 domains, we can generate 6 different domain adaptation tasks. \textbf{Office-Home}~\cite{venkateswara2017deep} is a more challenging dataset for visual domain adaptation, consisting of 15,500 images from 65 classes in 4 domains: Artistic (Ar), Clipart (Cl), Product (Pr) and Real-World (Rw). Similarly, we can generate 12 different domain adaptation tasks by permuting the 4 domains. \textbf{COVID-19}~\cite{Zhang2020COVIDDADD} is a cross-domain medical image analysis dataset for the diagnosis of COVID-19, which consists of 11663 images from 3 classes in two domains. The source domain consists of normal cases and pneumonia cases, while the target domain consists of normal cases and COVID-19 cases. We follow the studies in~\cite{Zhang2020COVIDDADD,zhang2020collaborative} to transfer knowledge in identifying pneumonia cases to identify COVID-19 cases.

Since these three datasets are clean, we create their corrupted versions following the protocol in ~\cite{shu2019transferable, zhang2020collaborative}. In particular, we create noisy source data from the original clean data in three different ways: label corruption, feature corruption and mixed corruption. For label corruption, we change the label of each image uniformly to a random class with probability $p_{noise}$. For feature corruption, each image is corrupted by Gaussian blur and Salt-and-Pepper noise with probability $p_{noise}$. As for mixed corruption, each image is processed by label corruption and feature corruption with probability $p_{noise}/2$ independently. We term $p_{noise}$ as the noise level for a domain adaptation task. In all the experiments, we use the noisy data for the source domain and clean data for the target domain. 

\textbf{Bing-Caltech}~\cite{bergamo2010exploiting} dataset contains Bing dataset and Caltech-256 dataset. The Bing dataset consists of images retrieved by Bing image search for each of the Caltech-256 category. Apart from the statistical differences between Bing images and Caltech images, the Bing dataset contains rich noise, with multiple objects in the same image. We use the Bing dataset as the noisy source domain and Caltech-256 as the clean target domain. While the experiments on Office-31, Office-Home and COVID-19 use manually synthesised noise, the experiments on Bing-Caltech report the performance for the real world weakly supervised domain adaptation.

\textbf{Implementation Details.} We adopt the 50-layer ResNet~\cite{he2016deep} as the feature extractor for all experiments on general images and MobileNet-V2~\cite{sandler2018mobilenetv2} as the feature extractor for all experiments on medical images. The hyper-parameter $T$ is set to 10, $\lambda_1$ is set to 0.5, $\lambda_2$ is set to 1, $\eta$ is set to 0.5. We employ SGD with weight decay 5e-4 to train the network. All experiments are repeated three times and we report the average result. For fair comparison, we report baseline results directly from the original papers if the experiment setting is the same and re-implement the methods to follow our experiment setting when there is difference.

\textbf{Baseline Methods.} We compare our method with state-of-the-art deep learning methods, domain adaptation methods, and weakly supervised domain adaptation methods. (1) \textbf{ResNet-50}~\cite{he2016deep} applies ResNet-50 trained on the noisy source domain to classify target data. (2) \textbf{MobileNet-V2}~\cite{sandler2018mobilenetv2} applies MobileNet-V2 trained on the noisy source domain to classify the target data. (3) \textbf{DANN}~\cite{ganin2014unsupervised} learns domain invariant features adversarially with gradient reversal. (4) \textbf{DAN}~\cite{long2015learning} applies multiple variants of MMD to align feature representations from multiple layers. (5) \textbf{RTN}~\cite{long2016unsupervised} extends DAN by adapting classifiers through a residual transfer module. (6) \textbf{ADDA}~\cite{tzeng2017adversarial} adopts asymmetric feature extractors for adversarial training. (7) \textbf{CDAN+E}~\cite{long2018conditional} conditions the adversarial adaptation models on discriminative information conveyed in the classifier predictions and uses the entropy of prediction as an importance weight. (8) \textbf{CLAN}~\cite{luo2019taking} conducts category-level domain adaptation. (9) \textbf{TCL}~\cite{shu2019transferable} learns a transferable curriculum for WSDA. (10) \textbf{CoUDA}~\cite{zhang2020collaborative} performs collaborative unsupervised domain adaptation for WSDA. 

\begin{figure*}[t]
\includegraphics[width=0.95\linewidth]{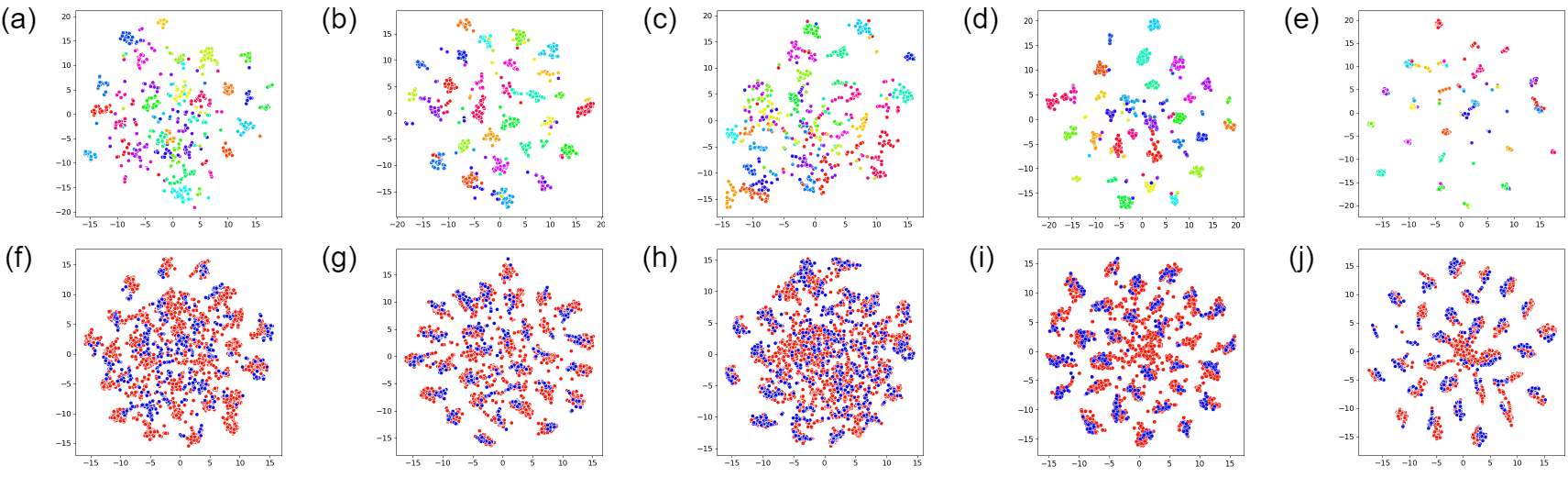}
\caption{The t-SNE visualization of DANN, CDAN+E, TCL, CoUDA and NTDA with class labels (a)-(e) for target data and domain labels (f)-(j) for both source and target data in the embedding space. In (a)-(e), different colors represents different classes. In (f)-(j), red color represents source data and blue color represents target data.}
\label{fig3}
\vspace{-2mm}
\end{figure*}

\subsection{Experimental Analysis}
\textbf{Performance Comparison on General Images.} We present the results on Office-31 under 40\% label corruption, feature corruption and mixed corruption in Table~\ref{table:1} and the results on Office-Home under 40\% mixed corruption and Bing-Caltech in Table~\ref{table:2}. Our method outperforms all the baseline methods in almost all tasks and significantly pushes forward the state-of-the-art performance on Office-31 by improving the average accuracy by 6.8\%, 2.4\%, 2.9\% on the label corruption, feature corruption and mixed corruption tasks respectively compared to the second best. It improves the average accuracy on Office-Home by 5.2\%, and the accuracy on Bing-Caltech by 1.3\% compared to the second best. For some tasks, the improvement of accuracy is more than 10\%. More specifically, NTDA performs better than state-of-the-art WSDA methods TCL~\cite{shu2019transferable} and CoUDA~\cite{zhang2020collaborative}, indicating the effectiveness of our method for transferring knowledge from noisy source domain to unlabeled target domain. NTDA also outperforms state-of-the-art UDA methods by a large margin, indicating that existing UDA methods indeed suffer from the noise in source domain, thus it is necessary to come up with methods to counter the negative effects of noisy data.

\begin{table} [hb]
\begin{center}
\resizebox{0.65\linewidth}{!}{%
\begin{tabular}{l|cccc}
\hline
Method &Acc (\%)&MP&MR&Macro F1\\
\hline
MobileNet-V2~\cite{sandler2018mobilenetv2}&58.5&47.8&31.3&37.1 \\
DANN~\cite{ganin2014unsupervised}&87.2&51.6&55.6&51.1 \\
CLAN~\cite{luo2019taking}&95.6&75.0&71.1&72.9 \\
\hline
TCL~\cite{shu2019transferable}&97.2&86.9&78.6&82.2 \\
CoUDA~\cite{zhang2020collaborative}&98.1&94.3&82.4&87.3 \\
\hline
NTDA&\textbf{98.3}&\textbf{99.3}&\textbf{89.2}&\textbf{93.6} \\
\hline
\end{tabular}}
\end{center}
\caption{Comparison on \textbf{COVID-19} diagnosis with 10\% label corruption.}
\label{table:covid}
\vspace{-4mm}
\end{table}

\textbf{Performance Comparison on Medical Images.} Following \cite{zhang2020collaborative}, we apply 10\% of label corruption on source domain of COVID-19 dataset and present the results in Table~\ref{table:covid}. We use Accuracy (Acc), Macro Precision (MP), Macro Recall (MP) and Macro F1-measure as metrics. Our method significantly outperform baseline methods in all metrics. More specifically, our method, NTDA,  outperforms CoUDA~\cite{zhang2020collaborative}, which is state-of-the-art method for the task. The results demonstrate the applicability of our method for cross-domain medical image analysis. In viewing that medical annotations are subjective and contain noises for domain adaptation, NTDA offers an appealing solution to the problem.

\begin{table} [ht]
\begin{center}
\resizebox{\linewidth}{!}{%
\begin{tabular}{l|ccccccc}
\hline
Method &A$\rightarrow$W&W$\rightarrow$A&A$\rightarrow$D&D$\rightarrow$A&W$\rightarrow$D&D$\rightarrow$W&Avg\\
\hline
NTDA (W/o UNR, CAA)&57.1&46.5&64.1&51.0&89.2&77.4&64.2 \\
NTDA (W/o CAA)&74.4&62.9&78.5&57.8&97.4&90.9&74.2 \\
NTDA (W/o UNR)&84.7&64.9&\textbf{88.3}&67.4&97.6&92.2&82.5 \\
NTDA&\textbf{89.5}&\textbf{68.3}&87.1&\textbf{68.8}&\textbf{98.7}&\textbf{95.1}&\textbf{84.6} \\
\hline
\end{tabular}}
\end{center}
\caption{Classification Accuracy (\%) on \textbf{Office-31} with 40\% mixed corruption.}
\label{table:3}
\vspace{-4mm}
\end{table}

\textbf{Ablation Study.}
Table~\ref{table:3} shows the performance of different variants of our model on Office-31 with 40\% mixed corruption. NTDA (W/o UNR, CAA) is the variant without Unsupervised Noise Removal and Cluster-Level Adversarial Adaptation. NTDA (W/o CAA) is the variant without Cluster-Level Adversarial Adaptation. NTDA (W/o UNR) is the variant without Unsupervised Noise Removal. Removing noisy source data and reducing the distribution shift across domains can both boost the target performance significantly, improving the accuracy by 10\%, 18.3\% respectively. NTDA combines unsupervised noise removal and cluster-level adversarial adaptation to achieve the best performance. Note NTDA (W/o UNR) also shows quite good performance, which is better than state-of-the-art WSDA methods, TCL~\cite{shu2019transferable} and CoUDA~\cite{zhang2020collaborative} on the same task. This is because our CAA method by itself is also robust to source data noise to some extent.

\begin{figure}[ht]
    \centering
    \includegraphics[width=0.5\linewidth]{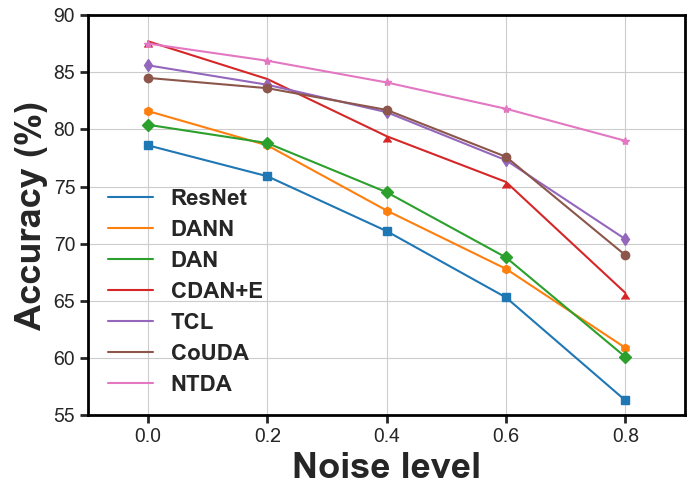}
\caption{Classification Accuracy (\%) w.r.t noise level.}
\label{fig4}
\vspace{-4mm}
\end{figure}

\textbf{Noisy Level.}
Fig.~\ref{fig4} shows the average classification results on Office-31 with mixed corruption under various noise levels. With the increase of noise level, the performance of all comparison methods degrade rapidly, while NTDA is more stable and provides much better performance which indicates that our method can handle various scenarios under weakly-supervised domain adaptation. In addition, NTDA performs as well as state-of-the-art domain adaptation method CDAN+E~\cite{long2018conditional} even when noise level is 0, indicating that our method is also applicable in standard domain adaptation scenario.

\begin{table} [ht]
\begin{center}
\resizebox{\linewidth}{!}{%
\begin{tabular}{l|cccccccc}
\hline
\multicolumn{2}{c|}{Method}&A$\rightarrow$W&W$\rightarrow$A&A$\rightarrow$D&D$\rightarrow$A&W$\rightarrow$D&D$\rightarrow$W&Avg\\
\hline
\multirow{2}{*}{P(\%)} & \multicolumn{1}{c|}{TCL~\cite{shu2019transferable}} & \multicolumn{1}{c}{96.8} & \multicolumn{1}{c}{86.3} & \multicolumn{1}{c}{96.8}& \multicolumn{1}{c}{82.4}& \multicolumn{1}{c}{86.5}& \multicolumn{1}{c}{84.3}& \multicolumn{1}{c}{88.8} \\
& \multicolumn{1}{c|}{NTDA} & \multicolumn{1}{c}{\textbf{99.2}} & \multicolumn{1}{c}{\textbf{96.2}} & \multicolumn{1}{c}{\textbf{98.2}}& \multicolumn{1}{c}{\textbf{95.3}}& \multicolumn{1}{c}{\textbf{96.3}}& \multicolumn{1}{c}{\textbf{96.6}}& \multicolumn{1}{c}{\textbf{97.0}} \\
\hline
\multirow{2}{*}{R(\%)} & \multicolumn{1}{c|}{TCL~\cite{shu2019transferable}} & \multicolumn{1}{c}{84.4} & \multicolumn{1}{c}{95.9} & \multicolumn{1}{c}{83.5}& \multicolumn{1}{c}{97.3}& \multicolumn{1}{c}{96.4}& \multicolumn{1}{c}{97.2}& \multicolumn{1}{c}{92.5} \\
& \multicolumn{1}{c|}{NTDA} & \multicolumn{1}{c}{\textbf{92.3}} & \multicolumn{1}{c}{\textbf{98.2}} & \multicolumn{1}{c}{\textbf{94.3}}& \multicolumn{1}{c}{\textbf{99.6}}& \multicolumn{1}{c}{\textbf{98.2}}& \multicolumn{1}{c}{\textbf{100.0}}& \multicolumn{1}{c}{\textbf{97.1}} \\
\hline
\end{tabular}}
\end{center}
\caption{Precision (P) and Recall (R) on \textbf{Office-31} with 40\% mixed corruption.}
\label{table:4}
\vspace{-6mm}
\end{table}

\textbf{Unsupervised Noise Modeling Quality.}
To test how well our unsupervised noise modeling method selects clean data, we compare it against the transferable curriculum in~\cite{shu2019transferable}. We use precision and recall as metrics for the comparison. Precision measures the fraction of clean data among the selected data, while recall measures the fraction of total number of clean data that are actually selected. Table~\ref{table:4} presents the precision and recall values of transferable curriculum and our method on Office-31 with 40\% mixed corruption. In all tasks, NTDA provide larger precision and recall results than transferable curriculum. On average, NTDA can select 97.1\% of clean data for training the network, and among the selected data only 3\% are noisy data which are significantly better than transferable curriculum which can only select 92.5\% of clean data, and among the selected data there are 11.2\% noisy data. Our method can select almost all clean data and remove all noisy data for training.

\textbf{Feature Visualization.}
Fig.~\ref{fig3} presents the t-SNE visualization of feature embeddings of DANN, CDAN+E, TCL, CoUDA and NTDA on task A$\rightarrow$W with 40\% mixed corruption. Fig.~\ref{fig3} (a)-(e) show the target feature embeddings by classes. NTDA model's embeddings are more compact and discriminative, while the rests' embeddings scatter and mix up among classes. Fig.~\ref{fig3} (f)-(j) show both the source and target feature embeddings by domain. NTDA aligns the target feature very well with the source data, while the other methods align the target feature not so well and mismatch the decision boundaries between the two domains. For DANN and CDAN+E, noisy source data degrades their feature embeddings. For TCL and CoUDA, as they align marginal distribution across domains without considering the fine-grained semantic structure in the embedding space, their feature embeddings surfer from class misalignment problem. These results validate the effectiveness of our method.

\begin{figure}[ht]
  \includegraphics[width=\linewidth]{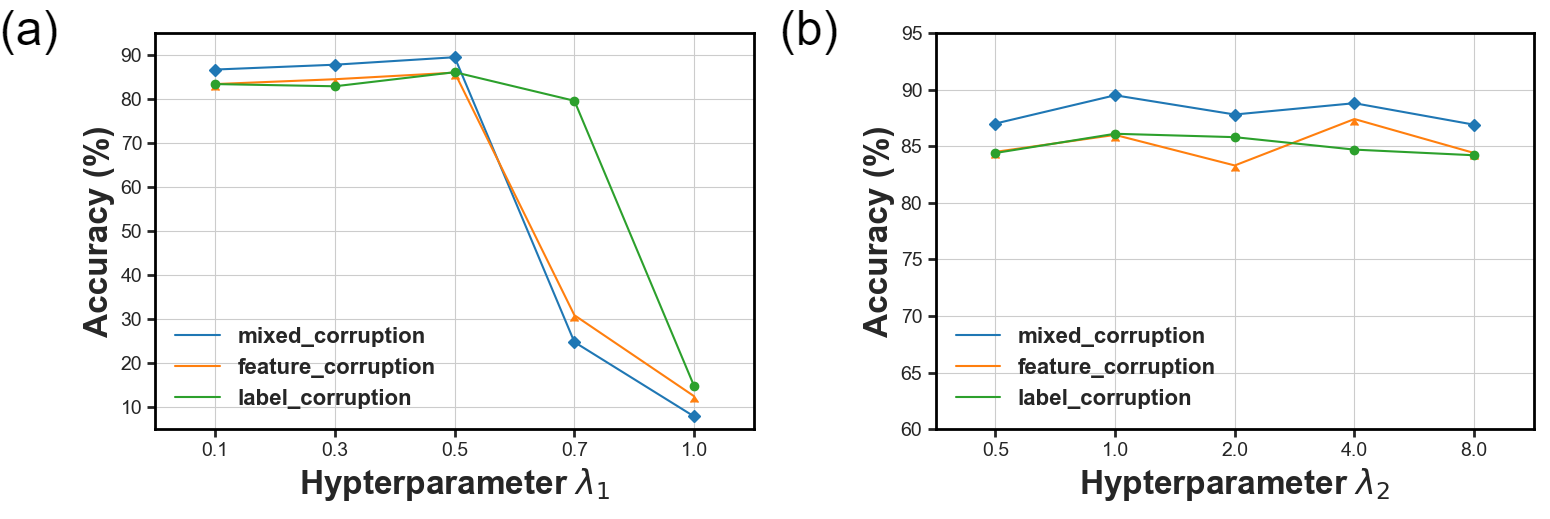}
\caption{Classification Accuracy (\%) on $A\rightarrow W$ under 40\% mixed corruption w.r.t (a) $\lambda_1$ and (b) $\lambda_2$.}
\label{fig6}
\end{figure}

\textbf{Hyper-parameters Sensitivity.}
Fig.~\ref{fig6}(a) shows the sensitivity analysis of NTDA on hyper-parameter $\lambda_1$ under 40\% label corruption, feature corruption and mixed corruptions. In general, NTDA performs stably when $\lambda_1$ is small, i.e. smaller than 0.5. When $\lambda_1$ becomes larger than 0.5, the performance of our method degrades. This is because $\lambda_1$ controls the strength of the compact regularizer to minimize the distances between data points and their class prototypes, when $\lambda_1$ becomes too large, it will tend to map all data points and class prototypes into a single point in the embedding space, thus making the performance much worse. Fig.~\ref{fig6}(b) shows the sensitivity analysis of NTDA on hyper-parameter $\lambda_2$ under 40\% label corruption, feature corruption and mixed corruption. In general, NTDA performs stably with the changes of $\lambda_2$. When $\lambda_2$ is within the interval $[0.5, 8]$, for label corruption and mixed corruption, the change of accuracy values is within 2\% and for feature corruption, the change of accuracy is within 5\%.

\begin{figure}[ht]
    \centering
    \includegraphics[width=0.8\linewidth]{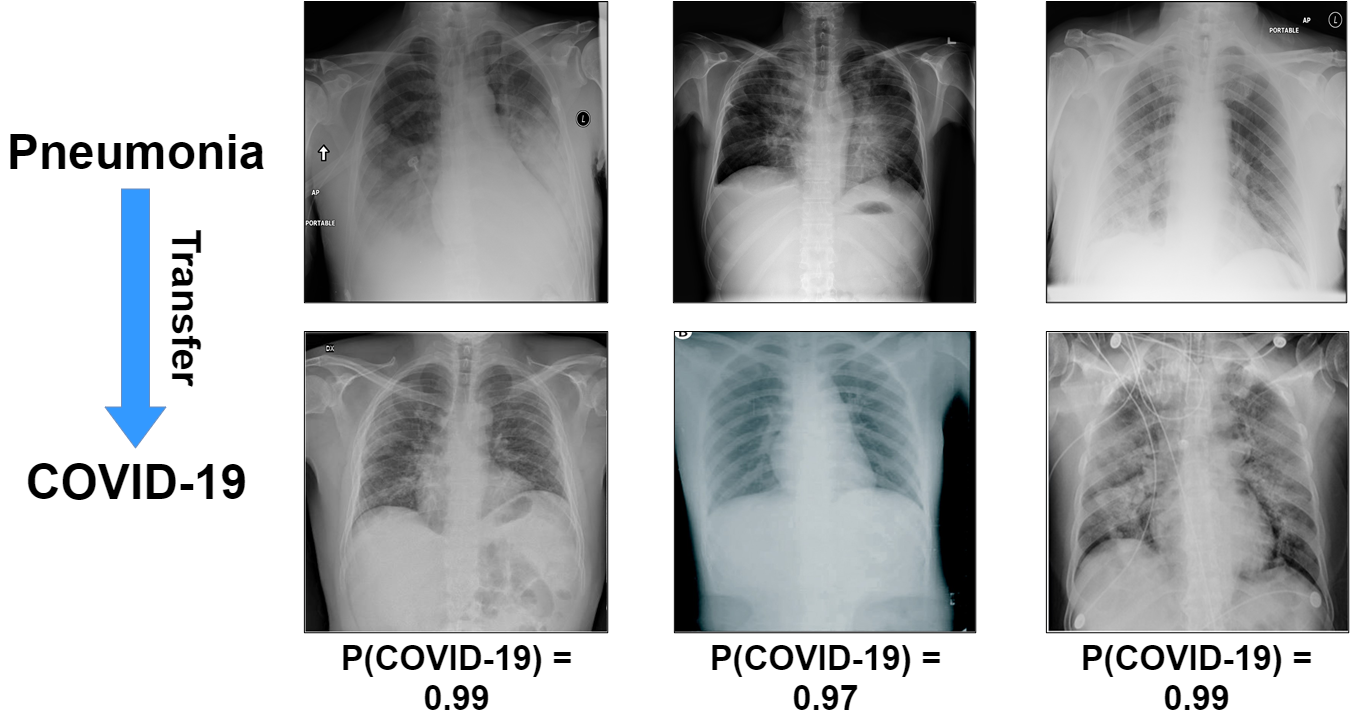}
\caption{Samples analysis of NTDA for cross-domain COVID-19 diagnosis under 10\% label corruption.}
\label{fig8}
\end{figure}

\textbf{Sample Analysis.} Fig.~\ref{fig8} shows the sample analysis of NTDA for the task of cross-domain COVID-19 diagnosis under 10\% label corruption. Although the task of identifying pneumonia cases is different from the task of identifying COVID-19 cases. Pneumonia cases share some similar characteristics with COVID-19 cases~\cite{zhang2020collaborative}. But there also exists distribution difference between the two datasets, e.g., image color difference and different artifacts displayed in the images. Despite these differences and the fact that the annotations in the source data are noisy, NTDA can still provide high prediction probabilities for target images on the correct class, which demonstrates the effectiveness of our method when applied for medical image analysis.

\section{Conclusions} \label{sec:conclusions}
In this paper, we study the under-explored Weakly Supervised Domain Adaptation problem (WSDA) for multimedia analysis. WSDA is a promising research area in view of its benefit to significantly reduce the annotation cost for deep learning. We identify several issues of existing WSDA methods and propose NTDA, a novel and effective method with unsupervised noise removal and cluster-level adversarial adaptation to alleviate the adverse effect of noisy data during domain adaptation. We conduct extensive experimental evaluation using four public datasets covering both general image analysis and medical image analysis, and the results show that our new method significantly outperforms existing methods. 

\begin{acks}
We thank the anonymous reviewers for their constructive comments, NUS colleagues and Beng Chin Ooi for their comments and contributions.
This research is supported by Singapore Ministry of Education Academic Research Fund Tier 3 under MOE's official grant number MOE2017-T3-1-007. Meihui Zhang's work is supported by the National Natural Science Foundation of China (62050099).
\end{acks}

\newpage
\bibliographystyle{ACM-Reference-Format}
\bibliography{sample-base}


\end{document}